\documentclass[journal]{IEEEtran} % use the `journal` option for SIPAIM conference style
\IEEEoverridecommandlockouts
\usepackage{cite}

\usepackage[pdftex]{graphicx}

\usepackage{amsmath}

\usepackage{booktabs}

\usepackage{algorithmic}

\usepackage{array}

\usepackage{xcolor}        % modern, preferred

\usepackage{placeins}   % gives \FloatBarrier

\renewcommand{\textcolor}[2]{#2} 
\usepackage{url}
\usepackage{hyperref}
\def\BibTeX{{\rm B\kern-.05em{\sc i\kern-.025em b}\kern-.08em
    T\kern-.1667em\lower.7ex\hbox{E}\kern-.125emX}}

\begin{document}


%use previous title.
\title{FedDAPL: Toward Client-Private Generalization in Federated Learning}
\author{%%%% author names
    \IEEEauthorblockN{1\textsuperscript{st} Soroosh Safari Loaliyan}% first author
    , \IEEEauthorblockN{2\textsuperscript{nd} Jose-Luis Ambite}% delete this line if not needed
    , \IEEEauthorblockN{3\textsuperscript{rd} Paul M. Thompson}% delete this line if not needed
    , \IEEEauthorblockN{4\textsuperscript{th} Neda Jahanshad}% delete this line if not needed
    , \IEEEauthorblockN{5\textsuperscript{th} Greg Ver Steeg}% delete this line if not needed
    % duplicate the line above as many times as needed to list all authors
    \\%%%% author affiliations
    \IEEEauthorblockA{\textit{Computer Science and Engineering (UCR), Riverside, US}}\\% first affiliation
    \IEEEauthorblockA{\textit{University of Southern California, Los Angeles, US}}\\% delete this line if not needed
    % duplicate the line above as many times as needed to list all affiliations
    %%%% corresponding author contact details
    \IEEEauthorblockA{soroosh.safariloaliyan@ucr.edu}
}

% GV suggested order: Soroosh, Jose-Luis Ambite, Paul Thompson, Neda Jahanshad, Greg Ver Steeg

\maketitle

% \begin{abstract}
%     % This document is a model and instructions for \LaTeX.
%     % This and the IEEEtran.cls file define the components of your paper [title, text, heads, etc.]. *IMPORTANT: Do Not Use Symbols, Special Characters, Footnotes, or Math in Paper Title or Abstract.

% \end{abstract}
\begin{abstract}
Federated Learning (FL) trains models locally at each research center or clinic and aggregates only model updates, making it a natural fit for medical imaging, where strict privacy laws forbid raw data sharing. A major obstacle is \emph{scanner‐induced domain shift}: non-biological variations in hardware or acquisition protocols can cause models to fail on external sites.  Most harmonization methods correct this shift by directly comparing data across sites, conflicting with FL’s privacy constraints.  Domain Generalization (DG) offers a privacy-friendly alternative—learning site-invariant representations without sharing raw data—but standard DG pipelines still assume centralized access to multi-site data, again violating FL’s guarantees.
This paper meets these difficulties with a straightforward integration of a Domain-Adversarial Neural Network (DANN) within the FL process. After demonstrating that a naive federated DANN fails to converge, we propose a proximal regularization method that stabilizes adversarial training among clients.  Experiments on T1-weighted 3-D brain MRIs from the OpenBHB dataset, performing brain-age prediction on participants aged \mbox{6–64\,y} (mean 22±6\,y; 45\% male) in training and \mbox{6–79\,y} (mean 19±13\,y; 55\% male) in validation, show that training on 15 sites and testing on 19 unseen sites yields superior cross-site generalization over FedAvg and ERM while preserving data privacy.

% \begin{abstract}
% We address these challenges with a simple framework that embeds a Domain-Adversarial Neural Network (DANN) inside the FL loop. After showing that a naïve federated DANN fails to converge, we introduce a proximal regularization technique that stabilizes adversarial training across clients. We evaluate on T1-weighted 3-D brain MRIs from the OpenBHB collection, performing brain-age prediction on participants aged 6–64 y (mean 22 y; 45 \% male) in training and 6–79 y (mean 19 y; 55 \% male) in validation. Training on 15 sites and testing on 19 unseen sites, our method surpasses federated baselines such as FedAvg and ERM, delivering stronger cross-site generalization while preserving data privacy.

\end{abstract}

\section{Introduction}

% Machine Learning (ML) has become an essential tool across various domains due to its ability to uncover intricate patterns from large datasets. Its applications span diverse fields such as cybersecurity~\cite{}, environmental monitoring and air quality prediction~\cite{}, healthcare diagnostics using medical imaging~\cite{}, and biomedical signal processing~\cite{}. Particularly, medical imaging has benefited significantly from ML advancements, enhancing diagnostic accuracy and efficiency in clinical workflows~\cite{}.

% Despite these successes, 
Machine Learning (ML) models often degrade when data from different sites or acquisition protocols are not harmonized properly. Lack of harmonization occurs when a model trained on a certain set of sources encounters new data with different statistical properties—e.g., scanner parameters, population demographics, or preprocessing workflows—leading to performance degradation~\cite{474d1aa2a24e412c84c797e66eb85e21}. Harmonization plays an essential part in fields such as medical diagnostics, where models need to be not only trustworthy but also able to generalize to new settings~\cite{ZUO2021118569}.

%10.1093/biostatistics/kxj037
To address domain shift issues, researchers have proposed several strategies, such as dataset harmonization~\cite{ FORTIN2017149, FORTIN2018104}, Domain Adaptation (DA)\cite{Guan_2022}, and Domain Generalization (DG)\cite{Yoon_2024}. Dataset harmonization matches statistical distributions between datasets, usually via style transfer~\cite{Liu2023StyleTransfer} or statistical normalization techniques~\cite{FORTIN2018104}. DA needs some access to the target domain and adjusts the source model accordingly~\cite{safaei2025certaintyuncertaintyguidedactive}. In contrast, DG seeks to train a model that not only works well over the training domains but also has high accuracy—without additional adaptation—on entirely unseen test domains~\cite{10.1007/s10994-009-5152-4}.

% Explain Doamin generalization

Domain Generalization has shown great potential in medical imaging tasks~\cite{Ghaffari2025.02.20.637919}, including MRI analysis, improving the robustness of brain lesion segmentation~\cite{Kamraoui2022_DLB}, Alzheimer's disease detection~\cite{Batool2025_SDGAD}, and amyloid-positivity classification~\cite{Li2022_DG_Amyloid}. In particular, brain age prediction from 3D MRI scans is a very common biomarker used for monitoring neurological aging and early disease detection. However, its practical deployment encounters problems such as differences in MRI acquisition protocols between institutions and the natural privacy limitations inherent in medical data.

Federated Learning (FL)\cite{li2024federateddomaingeneralizationsurvey, gargary2024systematic} has been a promising approach that allows collaborative training across various decentralized institutions without compromising data privacy~\cite{DBLP:journals/corr/McMahanMRA16, Plis2016COINSTAC, ghinani2025fuseflfullysecurescalable, Stripelis2024MetisFL}. Still, direct application of Domain-Adversarial Neural Networks (DANN)\cite{ganin2016domain}, a well-known DG method, to federated settings has limitations. In this work, we tackle these challenges by presenting FedDAPL, a federated domain-generalization framework that adds a proximal regularization term to effectively stabilize domain-adversarial training among federated clients.

Our research utilizes the OpenBHB dataset, which consists of roughly 4,000 MRI scans gathered from 70 distinct sites. Our models are trained on a subset of 1,587 images from 15 sites, which is significantly dominated by a single site (site 12), making up about 60\% of the data. This inherent bias in the training distribution makes domain shift more challenging during evaluation. We then validate our models on 19 entirely unseen sites to thoroughly test their generalization capabilities. %mention we are using DG, not harmonization

The main contributions of this paper include:
\begin{itemize}
\item Presenting FedDAPL, a new solution that successfully integrates  Federated Learning with Domain-Adversarial Neural Networks using proximal regularization to mitigate domain shift.
\item Investigating the impacts of proximal loss on domain generalization performance.
\item Performing extensive experiments showing our approach is superior to conventional benchmarks such as FedAvg, Empirical Risk Minimization (ERM), and naive federated DANN implementations.
\end{itemize}

\section{Background and Related Work}

% \subsection{Federated Learning}
\textbf{\textit{Federated Learning:}} 
FL allows multiple clients to collaboratively train models without directly sharing sensitive local data. In FL, all clients update their local model parameters individually, and the server combines these updates to create a global model. The most widely adopted FL algorithm is Federated Averaging (FedAvg)~\cite{DBLP:journals/corr/McMahanMRA16}, which aggregates client model parameters by simply averaging their updates every round. Simple as it is, FedAvg does not work well for non-IID data, motivating extensions such as FedProx~\cite{li2020federatedoptimizationheterogeneousnetworks}, SCAFFOLD~\cite{pmlr-v119-karimireddy20a}, and FedDAG~\cite{che2025feddagfederateddomainadversarial} to address heterogeneity across clients.

Among these techniques, FedProx is particularly relevant as it adds a proximal regularization term to stabilize training by penalizing deviations between local and global models. Related concepts are seen in FedHarmony~\cite{dinsdale2022fedharmonyunlearningscannerbias}, which integrates harmonization strategies with FL frameworks for reducing data distribution gaps between clients. Our approach likewise integrates a proximal loss into FedAvg for stabilizing adversarial learning between clients, albeit our application is specifically designed for federated domain generalization (DG).

% \subsection{Domain Generalization}
\textbf{\textit{Domain Generalization:}} 
Domain Generalization (DG) aims at building models robust enough to generalize well to unseen target domains by leveraging data from multiple source domains. Unlike domain adaptation, DG methods do not require access to the target domain data during training. Some popular DG methods include Domain-Adversarial Neural Networks (DANN)~\cite{ganin2016domain}, Invariant Risk Minimization (IRM)~\cite{arjovsky2020invariantriskminimization}, and DeepCORAL~\cite{sun2016deepcoralcorrelationalignment}.

Among these approaches, DANN is highly successful. It uses adversarial training to learn domain-invariant features by pushing the feature extractor to fool a domain discriminator. Through this, the model learns to generate generalized features that are less sensitive to domain shifts and thus is effective in a wide range of DG settings~\cite{ganin2016domain}.

Previous work has conducted a comprehensive evaluation of several DG methods—namely, DANN, IRM, ERM, DeepCORAL, DANNCE, and commonly-used baseline Empirical Risk Minimization (ERM)~\cite{NIPS2006_b1b0432c}—using the Stanford WILDS package~\cite{koh2021wildsbenchmarkinthewilddistribution, sagawa2022extendingwildsbenchmarkunsupervised} on the OpenBHB dataset, specifically targeting brain age prediction from 3D MRI data. Our results demonstrated that among these methods, DANN achieved the best out-of-domain generalization performance~\cite{10678307}.

% \subsection{Federated Domain Generalization and Our Prior Work}
\textbf{\textit{Federated Domain Generalization:}} 
Integrating federated learning with DG is a developing field aimed at overcoming both data privacy and domain shift issues. Methods like FedDG~\cite{liu2021feddgfederateddomaingeneralization}, FedDAG~\cite{che2025feddagfederateddomainadversarial}, and FedHarmony~\cite{dinsdale2022fedharmonyunlearningscannerbias} suggest federated adaptations of DG methods to handle heterogeneity and domain shifts without sacrificing privacy.

Inspired by these results, this paper investigates federated domain generalization through the proposal of FedDAPL, a federated learning method that combines FedAvg with a proximal-domain adversarial technique. In particular, we compare FedDAPL with multiple baselines, including FedAvg, a naive implementation of federated DANN without proximal regularization (Naive FedDANN), ERM, and centralized DANN, the latter usually acting as an upper-bound baseline. In terms of performance metrics, we mainly compare the Mean Absolute Error (MAE) on unseen validation domains (OOD). Additionally, as space allows, we also investigate the average L1 loss and study the sensitivity of our method to the proximal regularization hyperparameter ($\mu$). Our experimental results indicate FedDAPL outperforms FedAvg and Naive FedDANN significantly, achieving near-ERM performance while preserving privacy requirements.

\section{Methodology}

% In this section, we introduce our federated domain generalization framework, FedDAPL, specifically designed for robust MRI-based brain age prediction across diverse data acquisition sites. The model consists of three primary components: a \textbf{feature extractor}, a \textbf{regression head}, and a \textbf{domain discriminator}.
We train a neural network to predict a clinically relevant target, in this case, brain age, as a stand-in for more general diagnostic tasks. Training is done federatively across several sites so that raw MRI data never leave local institutions. A common strategy for achieving site invariance in adversarial training is to employ a discriminator that tries to guess site identity from intermediate features, while the feature extractor learns not to reveal that information —a tactic demonstrated to enhance cross-site generalization~\cite{ganin2016domain}. However, in a federated setting, though, each local discriminator sees data from only its site, which makes the classification task trivial and yields weak, uninformative gradients.

To overcome this, we introduce \textbf{FedDAPL}, which augments the typical prediction and adversarial losses with a proximal term that ties local feature updates to a common reference. \textcolor{blue}{The proposed FedDAPL framework is not a new network architecture, but an adaptation of Domain-Adversarial Neural Networks (DANN) to federated learning. Its contribution is a discriminator-specific proximal regularizer that anchors local adversaries to the global discriminator, preventing collapse into trivial site classifiers while letting the feature extractor and regressor adapt freely. This contrasts with FedProx, which applies the proximal penalty to \emph{all} parameters.} The model consists of a feature extractor, a regression head for brain age prediction, and a domain discriminator; FedDAPL simultaneously optimizes prediction loss, adversarial site loss, and proximal consistency loss. The detailed objectives and model specifications are specified below.

% \textbf{\textit{Feature extractor.}} Five consecutive 3D convolutional blocks; each block contains Conv3D, InstanceNorm3D, MaxPooling3D and ReLU.

% \textbf{\textit{Regressor.}} Two 3D convolutional blocks: the first has Conv3D, InstanceNorm3D, AveragePooling3D, and ReLU; the second is a Conv3D layer that outputs the scalar brain‑age prediction.

% \textbf{\textit{Domain discriminator.}} A fully connected network:
% Linear $\rightarrow$ BatchNorm $\rightarrow$ ReLU $\rightarrow$ Linear $\rightarrow$ BatchNorm $\rightarrow$ ReLU $\rightarrow$ Linear,
% with 512 hidden units in each hidden layer.

% \subsection{Federated Domain-Adversarial Training}

\textbf{\textit{Federated Domain-Adversarial Training.}}
This work extends the Domain-Adversarial Neural Network (DANN) to a federated learning environment. Both the feature extractor, regressor, and discriminator of each federated client are trained locally. Following local training, the central server aggregates updates via FedAvg on both the feature extractor/regressor and discriminator parameters.

Formally, the local training goal on client $k$ includes three loss terms: (1) prediction loss ($\mathcal{L}_y$), (2) domain classification loss ($\mathcal{L}_d$), and (3) proximal regularization loss ($\mathcal{L}_{prox}$).

In particular, our local client loss is as follows:
% \begin{equation}
% \mathcal{L}_k = \mathcal{L}_y + \mathcal{L}_d + \frac{\mu}{2}\|\theta_{d,k} - \theta_{d,global}\|^2
% \end{equation}
\begin{equation}
\mathcal{L}_k \;=\; \mathcal{L}_y \;+\; \mathcal{L}_d 
\;+\; \mathcal{L}_{\text{prox}}
\label{eq:local_loss}
\end{equation}
Explicitly, these losses are calculated as:
{\setlength{\jot}{2pt}%
\begin{align}
\mathcal{L}_y   &= \mathrm{MSE}(y_{\text{pred}}, y_{\text{true}}), \\
\mathcal{L}_d   &= \mathrm{CrossEntropy}(d_{\text{pred}}, d_{\text{true}}), \\
\mathcal{L}_{\text{prox}} &= \frac{\mu}{2}\,\bigl\lVert \theta_{d,k} - \theta_{d,\text{global}} \bigr\rVert_2^{2} \label{eq:prox}
\end{align}
}
where $\theta_{d,k}$ is the local discriminator parameters of client $k$, and $\theta_{d,global}$ is the global discriminator parameters obtained from the server at the start of every round~\eqref{eq:prox}. The hyperparameter $\mu$ controls the intensity of the proximal regularization.

% \begin{align}
% \mathcal{L}_y &= \text{MSE}(y_{pred}, y_{true}),\\
% \mathcal{L}_d &= \text{CrossEntropy}(domain_{pred}, domain_{true}),\\
% \mathcal{L}_{prox} &= \frac{\mu}{2}\sum_{p, p_g \in (\theta_{d,k}, \theta_{d,global})}(p - p_g)^2.
% \end{align}

% \subsection{Training and Evaluation}
\textcolor{blue}{\textbf{\textit{Dataset Details.}}
This work uses OpenBHB T1w 3D MRIs~\cite{DUFUMIER2022119637} with no additional preprocessing beyond the standard release.
Data are split 75\%/25\% into training/OOD validation.
Training has 1,587 scans from 15 sites (6–64\,y; $22\pm6$; 45\% male, 55\% female); 
validation has 594 scans from 19 unseen sites (6–79\,y; $19\pm13$; 55\% male, 45\% female).
Each client holds data from three distinct sites; no site overlaps across clients. 
To reflect a realistic cross-silo imbalance, one client contains $\sim60$\% of the training set. 
The OOD split intentionally includes more sites than training to stress cross-site generalization with fewer training domains.
}

\textbf{\textit{Training and Evaluation.}}
Here, we consider a federated setup with multiple sites, each of which has MRI scans as well as corresponding labels (here, brain age). The goal is to jointly train a prediction model without requiring raw data exchange. Concretely, FedDAPL uses FedAvg on $K$ clients ( $K=5$ in our experiment) for global rounds of $R=15$. In each round, each client updates its local model for $E=5$ epochs based on its own MRI images (including age and site labels), followed by the server's weight averaging. To provide stable estimates, we repeat the entire federated training procedure 10 times with different random seeds and report average performance metrics. For testing, we evaluate the final global model on a held‑out set of 594 scans drawn from 19 unseen sites to measure cross‑site generalization. For the centralized baselines (ERM and DANN), we train a single model for 75 epochs and validate every 5 epochs with the same protocol.

% \subsection{Model Architecture}
\textbf{\textit{Model Implementation \& hyperparameters.}}
All the models are trained with the Adam optimizer. In centralized ERM, the feature extractor and regressor are trained with a learning rate of $8\times 10^{-4}$. In centralized DANN, their learning rate is $8.5\times 10^{-4}$, and the discriminator is trained using $2\times 10^{-3}$. Label smoothing (0.06) and gradient clipping (10.0) are used for the discriminator. In Federated and Centralized DANN, the reversed gradient is multiplied by a time-varying coefficient that determines the strength of the GRL regularize:
\[
\lambda_{\text{GRL}}(p) = 8.5\!\left(\frac{2}{1 + e^{-7p}} - 1\right),
\quad p=\frac{\text{epoch}}{75},
\]
with a 10‑epoch warm‑up during which $\lambda_{\text{GRL}}{=}0$. Learning-rate decay for both the feature extractor/regressor in ERM and DANN is controlled using PyTorch’s \texttt{ReduceLROnPlateau} (mode = \texttt{'min'}). The identical optimizer setup is used in FedDAPL, with the proximal weight $\mu \in \{0,10,20,40,100\}$. FedDAPL utilizes the architecture model proposed in ~\cite{gupta2021improved} and also expanded in ~\cite{gupta2023transferring}. It employs the model from WILDS~\cite{koh2021wildsbenchmarkinthewilddistribution} as the discriminator architecture with a hidden layer of size 512.

\section{Experiments and Results}

\textbf{Setup.}  
All experiments use the \textit{OpenBHB} brain MRI dataset.  The centralized baselines (\textbf{ERM} and \textbf{DANN}) train for 75 epochs on combined source data, validating every 5 epochs.  For the federated setting, \textbf{FedAvg}, \textbf{Naive FedDANN}, and \textbf{FedDAPL} train for 15 communication rounds with 5 local epochs per client each.  All numbers are averaged across 10 independent seeds.

\textbf{Centralized reference point.}  
Table \ref{tab:centralized_mae} demonstrates that incorporating a domain-adversarial loss (DANN) with ERM reduces the mean-absolute-error (MAE; in years) on unseen sites by 10\,\%.  This improvement establishes the accuracy “headroom’’ we aim to recover when privacy constraints forbid data pooling.

\textbf{Federated baselines and the heterogeneity gap.}  
In FL, \textbf{FedAvg} performs significantly worse than centralized ERM—aligned with findings that client heterogeneity impedes FedAvg convergence \cite{yao2025taskagnosticfederatedlearning}. Adding a local discriminator on every client (\textbf{Naive FedDANN}) does not improve and tends to be unstable.

\textbf{FedDAPL closes the gap.}  
By adding a proximal regularizer that softly aligns discriminators, \textbf{FedDAPL} recovers the centralized DANN advantage without compromising privacy.  It reduces MAE by 22\,\% relative to FedAvg (Table \ref{tab:federated_mae}) and reduces error across all 19 unseen test sites.

\textbf{Proximal-weight sweep.}  
Table \ref{tab:mu_sweep} sweeps the proximal weight~$\mu$.  Performance improves from $\mu{=}0$ (the naive case) to an optimum $\mu{=}40$; higher values overly restrict training and decrease accuracy.  All reported results utilize $40$.

\textcolor{blue}{
\textbf{Relationship to FedProx.}  
To clarify novelty, we compared FedDAPL (proximal penalty only on the discriminator) with FedProx (proximal penalty on all parameters). 
Both Adam and SGD were tested under $\mu \in \{20,40,100\}$. 
As shown in Table~\ref{tab:federated_mae}, FedProx consistently underperforms FedDAPL, indicating that restricting the proximal penalty to the discriminator is more effective for stabilizing adversarial alignment.}

\textcolor{blue}{
\textbf{Comparison to prior OpenBHB results.}  
Following the default OpenBHB split used by Dufumier et al.~\cite{DUFUMIER2022119637} and Cheshmi et al.~\cite{cheshmi2023clusterfl}, 
We re-ran FedDAPL under the official protocol. 
FedDAPL achieves an MAE of $3.88$, compared to $4.14$ for FedAvg and $3.86$ for clustered FL reported in prior work, while FedProx reaches an MAE of $5.00$ under the same setting. 
Although our absolute MAE is higher due to broader inclusion of pediatric subjects and extended age range, this situates FedDAPL fairly within OpenBHB benchmarks.}
\begin{table}[htbp]
    \centering
    \vspace{-0.75em}
    \caption{Centralized validation MAE (lower is better).}
    \vspace{-0.5em} % Adjust this value as needed
    \label{tab:centralized_mae}
    \begin{tabular}{lcc}
        \toprule
        Method & MAE(Years) & $\Delta$ vs.\ ERM \\
        \midrule
        ERM  & $5.70\pm0.34$ & -- \\
        DANN & $5.12\pm0.21$ & $-0.58$ \\
        \bottomrule
    \end{tabular}
\end{table}
\textcolor{blue}{
\textbf{Impact of optimizer choice.} 
We also compared Adam and SGD+momentum as local optimizers while holding all other hyperparameters fixed. 
Table~\ref{tab:mu_sweep} shows that FedDAPL remains competitive under both optimizers, indicating robustness to optimizer choice. \emph{Note that the SGD configuration was not fine-tuned; we expect further improvements to be possible with a dedicated hyperparameter search.}}
\begin{table}[htbp]
\centering
\vspace{-0.75em}
\caption{Federated validation MAE (lower is better).}
\vspace{-0.5em}
\label{tab:federated_mae}
\begin{tabular}{@{}lccc@{}}
\toprule
Method & MAE (Years) & vs.\ ERM & vs.\ FedAvg \\
\midrule
FedAvg (Adam)               & $6.25\pm0.25$ & $+0.55$ & -- \\
Naive FedDANN (Adam)       & $7.28\pm1.19$ & $+1.58$ & $+1.03$ \\
\textcolor{blue}{FedProx (Adam, $\mu$=20)} & \textcolor{blue}{$9.33\pm0.24$} & \textcolor{blue}{$+3.63$} & \textcolor{blue}{$+3.08$} \\
\textcolor{blue}{FedProx (SGD, $\mu$=20)}  & \textcolor{blue}{$8.85\pm0.33$} & \textcolor{blue}{$+3.15$} & \textcolor{blue}{$+2.60$} \\
\textbf{FedDAPL (Adam)}  & \textbf{$5.62\pm0.34$} & \textbf{$-0.08$} & \textbf{$-1.66$} \\
\textcolor{blue}{\textbf{FedDAPL (SGD)}}   & \textcolor{blue}{\textbf{$6.32\pm0.28$}} & \textcolor{blue}{\textbf{$+0.62$}} & \textcolor{blue}{\textbf{$+0.07$}} \\
\bottomrule
\end{tabular}
\end{table}
\begin{table}[!htbp]
  \centering
  \vspace{-0.75em}
  \caption{Effect of proximal weight $\mu$ on the validation MAE (years).}
  \vspace{-0.5em}
  \label{tab:mu_sweep}
  \begin{tabular}{@{}lccccc@{}}
    \toprule
    Method & $\mu=20$ & \textbf{40} & 100 \\
    \midrule
    \textcolor{blue}{FedProx (Adam)}  & \textcolor{blue}{9.33} & \textcolor{blue}{9.27} & \textcolor{blue}{9.45} \\
    \textcolor{blue}{FedProx (SGD)}  & \textcolor{blue}{8.85} & \textcolor{blue}{9.27} & \textcolor{blue}{9.45} \\
    FedDAPL (Adam)  & 5.88 & \textbf{5.62} & 5.89 \\
    \textcolor{blue}{FedDAPL (SGD)}   & \textcolor{blue}{6.39} & \textcolor{blue}{6.32} & \textcolor{blue}{6.86} \\
    \bottomrule
  \end{tabular}
\end{table}
\section{Conclusion}
In the centralized environment, DANN is clearly superior to ERM. This accuracy is unmatched by standard FedAvg, as client heterogeneity corrupts the global model, and a naive federated DANN does not fix the problem. FedDAPL recovers the gains of adversarial domain alignment in a privacy-preserving environment by incorporating an efficient proximal regularizer and achieves competitive generalization across unseen medical imaging sites.
As a future direction, we will investigate approaches that eliminate the need for a separate discriminator and will use only the feature extractor and regressor weights to capture and mitigate site-specific differences, further simplifying the training process and improving model generalization.

% \subsection{Subsection heading}

% \paragraph{Subsubsection heading}
% Place figures and tables at the top and
% bottom of columns. Avoid placing them in the middle of columns. Large
% figures and tables may span across both columns. Figure captions should be
% below the figures; table heads should appear above the tables. Insert
% figures and tables after they are cited in the text. Use the abbreviation
% ``Fig.~\ref{fig}'', even at the beginning of a sentence.
%
%\begin{table}[htbp]
%    \caption{Table Type Styles}
%    \begin{center}
%        \begin{tabular}{|c|c|c|c|}
%            \hline
%            \textbf{Table} & \multicolumn{3}{|c|}{\textbf{Table Column Head}}                       %                                  \\
%            \cline{2-4}
%            \textbf{Head}  & \textbf{\textit{Table column subhead}}           & %\textbf{\textit{Subhead}} & \textbf{\textit{Subhead}} \\
%            \hline
%            copy           & More table copy$^{\mathrm{a}}$                   &                     %      &                           \\
%            \hline
%            \multicolumn{4}{l}{$^{\mathrm{a}}$Sample of a Table footnote.}
%        \end{tabular}
%        \label{tab1}
%   \end{center}
%\end{table}

%\begin{figure}[htbp]
%    \centerline{\includegraphics{fig1.png}}
%    \caption{Example of a figure caption.}
%    \label{fig}
%\end{figure}

\section*{Acknowledgment}
This work was supported in part by NIH grants R01AG081571 and R01AG087513.
% Acknowledgments goes here.

\ifCLASSOPTIONcaptionsoff
  \newpage
\fi

% \section*{References}

% Please number citations consecutively within brackets \cite{b1}. The
% sentence punctuation follows the bracket \cite{b2}. Refer simply to the reference
% number, as in \cite{b3}---do not use ``Ref. \cite{b3}'' or ``reference \cite{b3}'' except at
% the beginning of a sentence: ``Reference \cite{b3} was the first $\ldots$''
%
% Number footnotes separately in superscripts. Place the actual footnote at
% the bottom of the column in which it was cited. Do not put footnotes in the
% abstract or reference list. Use letters for table footnotes.
%
%Unless there are six authors or more give all authors' names; do not use
%``et al.''. Papers that have not been published, even if they have been
%submitted for publication, should be cited as ``unpublished'' \cite{b4}. Papers
%that have been accepted for publication should be cited as ``in press'' \cite{b5}.
%Capitalize only the first word in a paper title, except for proper nouns and
%element symbols.
%
%For papers published in translation journals, please give the English
%citation first, followed by the original foreign-language citation \cite{b6}.
\bibliographystyle{IEEEtran}
\bibliography{refs}

\end{document}